\documentclass{article}

\usepackage{setspace}

\usepackage{arxiv}
\usepackage{times}
\usepackage{helvet}
\usepackage{pgfplots}
\usepackage{tikz}
\usepackage{courier}
\usepackage{amsmath}
\usepackage{amssymb}
\usepackage{amsthm}
\usepackage{setspace}
\pgfplotsset{
    dirac/.style={
        mark=triangle*,
        mark options={scale=2},
        ycomb,
        scatter,
        visualization depends on={y/abs(y)-1 \as \sign},
        scatter/@pre marker code/.code={\scope[rotate=90*\sign,yshift=-2pt]}
    }
}

\begin{document}
%
\title{Differentiable Mask for Pruning Convolutional  and Recurrent Networks}

\author{Ramchalam Kinattinkara Ramakrishnan\thanks{Equal contribution, order decided by coin flip} \\ ramchalam.ramakrishnan@huawei.com \\ Huawei Noah's Ark Lab \And Eyy\"ub Sari\footnotemark[1] \\ eyyub.sari@huawei.com  \\ Huawei Noah's Ark Lab \And Vahid Partovi Nia\\ vahid.partovinia@huawei.com \\ Huawei Noah's Ark Lab}  
\maketitle
\begin{abstract}
Pruning is one of the most effective model reduction  techniques. Deep  networks require massive computation and such models need to be compressed to bring them on edge devices. Most existing pruning techniques are  focused  on vision-based models like convolutional  networks, while text-based models are still evolving. The emergence of multi-modal multi-task learning calls for a general method that works on vision and text architectures simultaneously.  We introduce a \emph{differentiable mask}, that  induces sparsity on various granularity  to fill this gap. We apply our method successfully to prune weights, filters, subnetwork of a convolutional architecture, as well as nodes of a recurrent network. 
\end{abstract}



%

\section{Introduction}
Recent models on machine translation, self-driving cars,  strategy games have shown game-changing breakthroughs. However, most of these models are highly over-parametrised for a  variety of reasons, ranging from the increase of computational  power to the lack of domain expertise. Subsequently, deploying these models on constrained edge devices is counter-intuitive. For instance, real-time updates to mobile phones is hampered by the model size. Consequently, the training and inference time are impacted. 
An alternative is to store deep models on the cloud rather than edge devices to overcome many of the edge implementation drawbacks, and  perform computation on the cloud server. However, the cons far outweigh the pros, especially in terms of security, and the latency in transferring the data to and from the cloud. Most of the models are preferred to be stored and computed on the edge in real applications.
This goal can be achieved only by simplifying neural networks computations. 

Many categories of simplifications include quantization \cite{Courbariaux_BinaryConnect_2015},  \cite{Hubaraetal_BinaryNeuralNet_2016}, low-rank compression, pruning, and network architecture search.  We propose  a  pruning technique while training a neural network, and induce structured sparsity through node, filter or subnetwork pruning. Given an input and an output node in a computational graph, we define a subnetwork as a subset of the graph that has a directed path from the input to the output node going through a series of transformations.

We introduce 
i) a simple technique of jointly pruning and training convolutional as well as recurrent networks.
ii) a differentiable mask function to allow rejuvenation of a pruned entity through the training process.
iii) additional regularizer to control the amount of pruned parameters directly.

\section{Related Work}
Pruning can be divided into major five categories, including 1. Mask-based, 2. Lasso/Group Lasso, 3. Magnitude-based, 4. Reinforcement Learning (RL), 5. Miscellaneous.

One of the earliest methods introduced \cite{Han_EfficientNN_2015} is a magnitude based approach where the weights with the lowest magnitude were considered to have low importance and hence removed. However, this technique inherently induces unstructured sparsity and at the inference level would require specialized hardware to make use of this sparsity. 

Various techniques have been used in masked-based approaches, such as using scaling factors as mask values \cite{Zhou_2018}. Others created  \emph{annealed pruning}, as a substitution to dropout,  and used non-trainable masks based on the weight magnitude \cite{Bartoldson_2018}. Alternatively, many other works were based on a similar mask-based approach \cite{Guo_DynamicSurgery_2016}. They use Straight Through Estimator (STE) for back-propagation for the mask function. Some other techniques make use of hard threshold of weights using a mask function that has similar drawbacks to magnitude based pruning.  \emph{Scalpel} was introduced  \cite{Scalpel_Yu_2017} for pruning based on \emph{SIMD hardware architecture}. Often in node pruning, a mask function is applied to prune nodes that are below a threshold value. They also use a STE in back-propagation. The STE has been shown to be effective in training quantized networks while retaining the accuracy.

\emph{Network Slimming}  \cite{NetworkSlimming_Liu_2017} has been the most widely adopted technique as a baseline and uses $\ell_1$ regularization. They impose $\ell_1$ regularization on the scaling factors of batch normalization layer and prune low-magnitude scaling factors. The technique is a post-training pruning method and the motivation is to identify insignificant output channels by pushing the batch normalization scaling factors to zero. \emph{Group lasso} \cite{Yuan_GroupLasso_2006} is also proposed  to control the amount of pruning at different levels \cite{Wen_2016}. The group lasso is applied on filters, channels, layers, and filters.

Channel-based pruning techniques include filter pruning using $\ell_1$ norm  \cite{Li_PruneFilters_2016}  as the saliency score. The relative importance of each filter in a layer is obtained by calculating the sum of absolute weights in the filter or the $\ell_1$ norm. This approach is equivalent to training a network with $\ell_1$ regularization and then pruning it. Unfortunately magnitude-based filter pruning does not correctly prune redundant filters, so \emph{Thinet} was introduced  \cite{Luo_ThiNetPrune_2017} to prune channels based on the output of the next layer. After pruning, the reconstruction loss is reduced by the use of scaling factors and the network is then fine-tuned. Even random pruning sometimes provides better results than $\ell_1$-norm magnitude pruning \cite{Mittal_2018}.

Some pruning methods benefit from the reinforcement learning formalism.  \cite{He_AMC_2018} make use of a deterministic policy gradient approach for pruning. The agent processes network  layer-wise, and the state-space consists of an encoding of various parameters. The agent outputs a compression ratio as an action (between 0 and 1) and the validation accuracy is used as reward to train the RL model. Once the sparsity ratio is obtained for each layer, the model is trained from scratch. The aim is to discover the exact number of non-redundant parameters in a layer for a network. The overall time taken for convergence of RL based approaches generally require large training time compared to other traditional techniques.

Other miscellaneous techniques include  \emph{optimal brain damage} \cite{Lecun_Optimal_1990}, that uses a second order Taylor expansion to evaluate the saliency of the parameters. The idea is to obtain  parameter saliency by observing the shift in the loss function as a result of its pruning. However, the use of magnitude-based pruning and the metric of importance of weight in the network is not theoretically sound as it does not take into account potential correlation between parameters. Another well-known method is discrimination-aware channel pruning  \cite{Zhuang_DisAware_2018} which is shown to be effective in compressing networks. They propose to start with  a pre-trained network and use strategies to obtain channels that truly contribute to the discriminative power of the network. They add extra discriminative losses to each convolutional layers and optimize the reconstruction error. \cite{NeuralRejuvenation_Qiao_2018} brought a novelty by rejuvenating the dead neurons while training. They use a binary mask to showcase the importance of each neuron in a layer. Once the neuron is rejuvenated, the weights are reinitialized or set to null. 

Methodologies mentioned above have pros and cons. Many of them use a pre-trained network and prune the network after training. This would require multiple iterations of pruning and fine tuning to improve the model. In magnitude-based pruned  weights with low magnitude correlate with other non-pruned weights, so the metric of importance of weights may not be sound. Most masked-based pruning techniques use a hard-threshold and a \emph{non-differentiable mask} and often fall on post-train pruning methods. Some methods use the $\ell_1$ norm of the filters which may not provide the best generalization capabilities and  need for a better metric for pruning. Having a low $\ell_1$ norm does not mean the filter is useless and correlation between the elements in the filter often plays a crucial role. 

The closest recent work to our technique is the sparsity induction through $\ell_1$ regularization \cite{huang_2018_data}. They make use of a scaling factor for pruning and push the scaling factors towards the origin to ensure sparsity. For optimization of the scaling factor, they used \emph{Accelerated Proximal Gradient}, which requires modification of the training process. Another comparable work \cite{diffpruning2019kim}  use a differentiable mask function that activates based on a latent parameter trained via back-propagation, and add  a compression ratio control regularizer. However, as they do not scale their output, the filters of the model could get deactivated abruptly and  consequently suffer from instability.

\begin{figure*}[ht]
  \centering
  \includegraphics[width=0.2\linewidth]{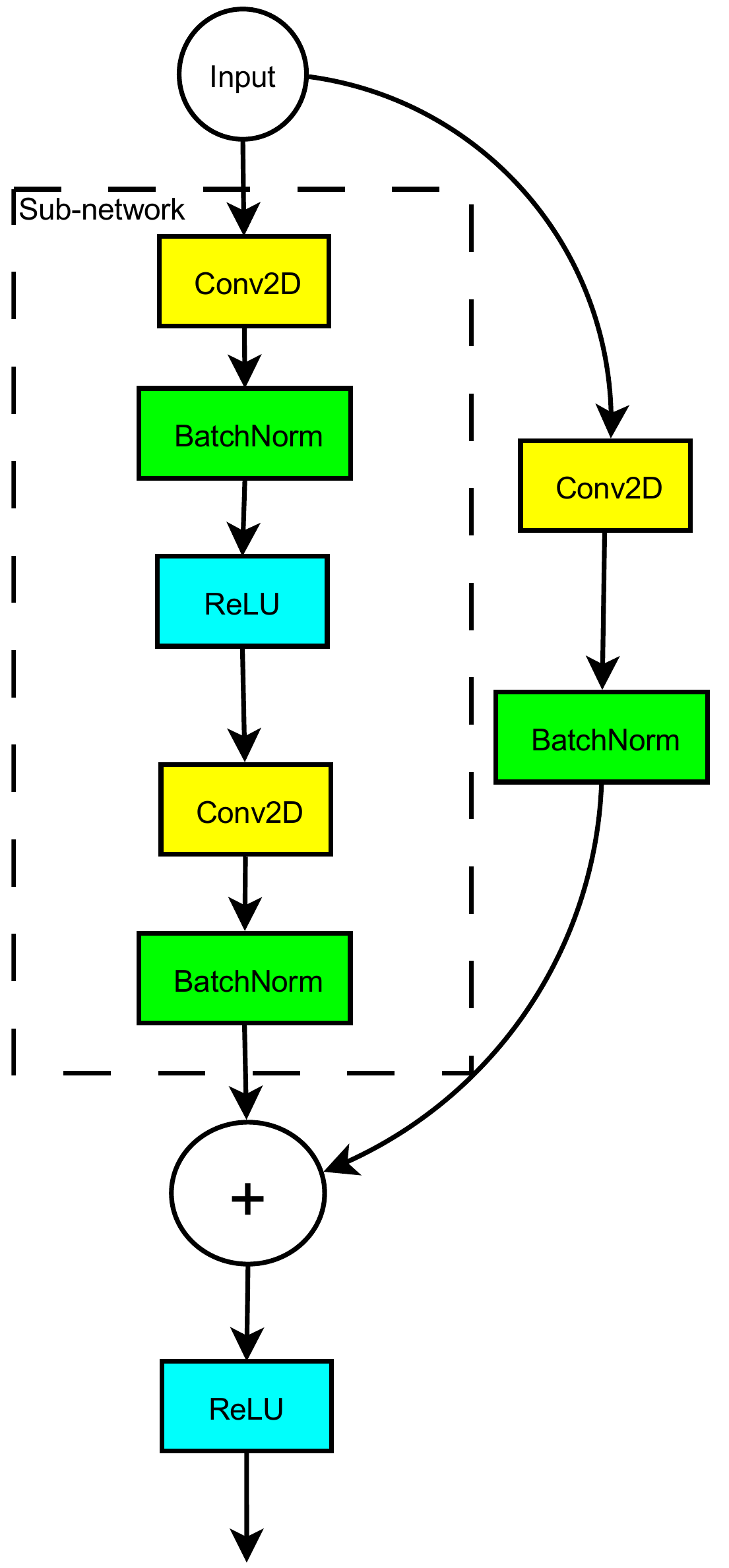}~~~~~
  \includegraphics[width=0.25\linewidth]{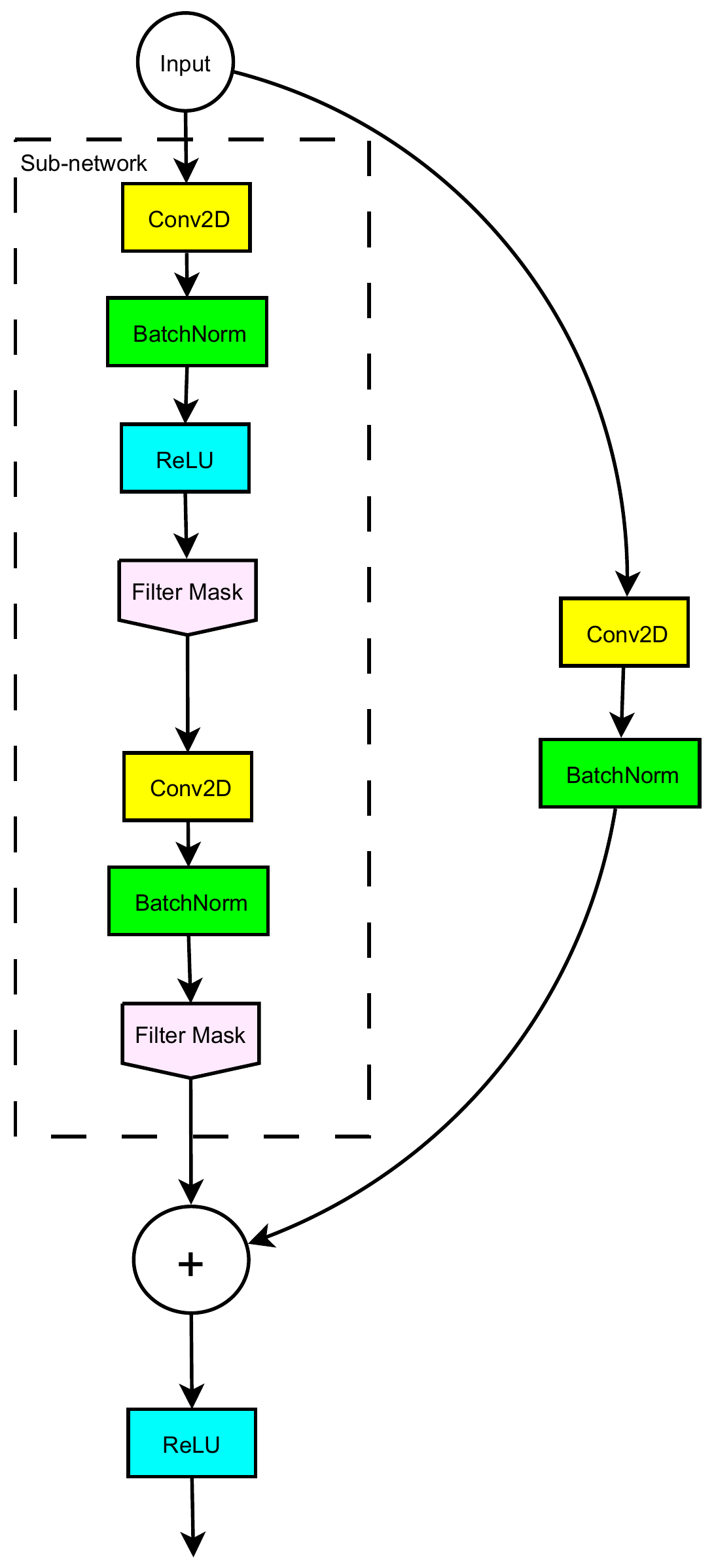}~~~
  \includegraphics[width=0.25\linewidth]{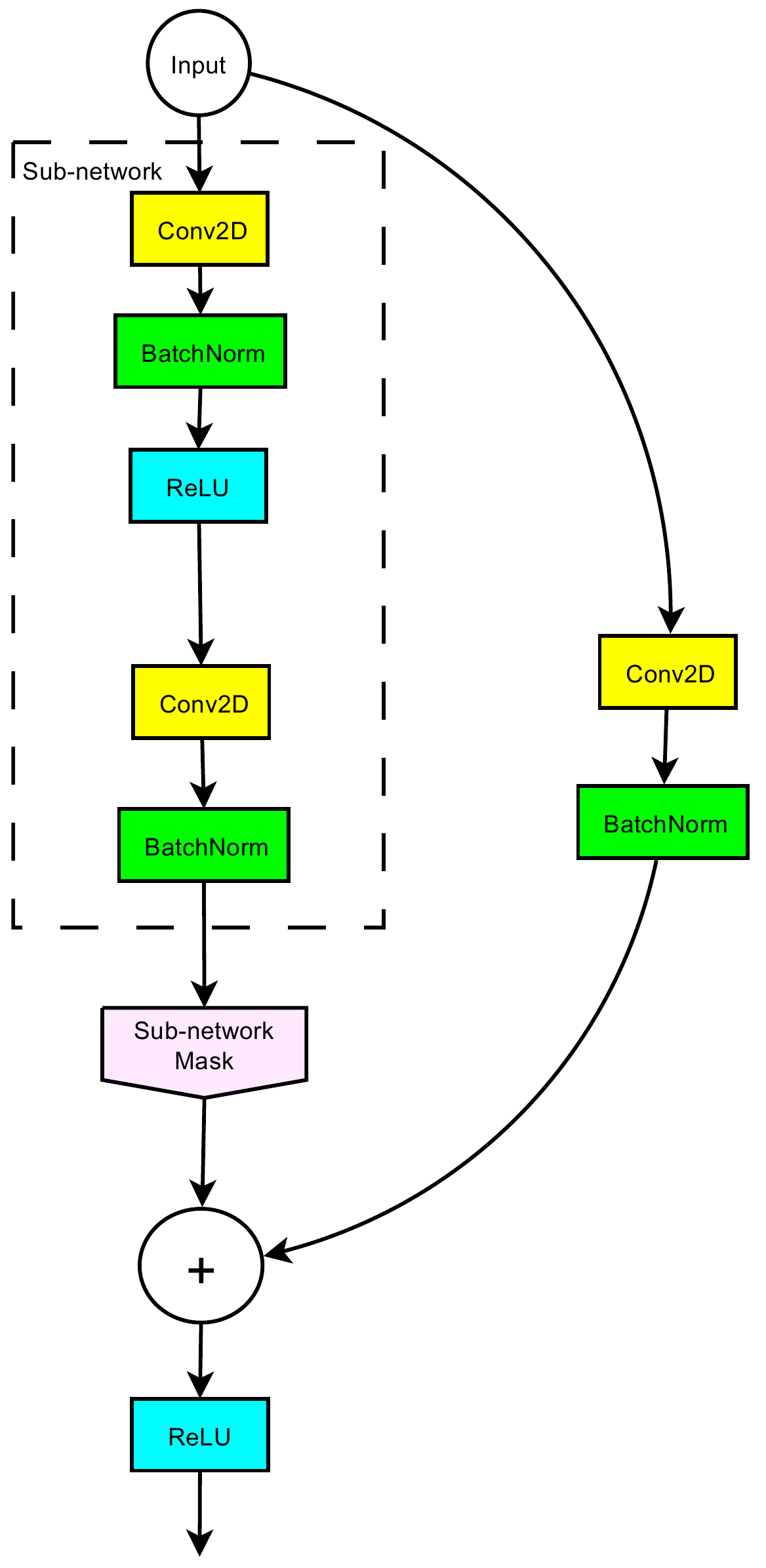}~~~
  \caption{A ResNet-style sub-network (left panel), DMP filter pruning (middle panel), DMP block pruning (right panel).}
  \label{fig:resnet_block}
\end{figure*}
\section{Proposed method}

Let $\{(\mathbf{x}_i, \mathbf{y}_i)~|~i\in N\}$ be a dataset of $N$ samples with $\mathbf x_i$ representing the input vector and $\mathbf y_i$ the output vector. We consider a model $M$ with $L$ layers, where a layer $l \in L$ represents a prunable entity and its parameters denoted by $\theta_l$. We define a prunable entity as a node in the computational graph that does not invalidate the graph upon its parameters being removed (i.e. the forward pass can still be performed). Let $f: \mathbb{R}^n \to \mathbb{R}^n$ be any element-wise transformation mapping  on a node's output (eg. ReLU, Batch Normalization, Identity etc.). Let $J(\theta)$ be the objective to be minimized, 
$$J^* = \min_{\theta\in \Theta}J(\theta),$$ 
where $\Theta$ is the space all learnable parameters and $J^*$ is the optimized loss.

We introduce \emph{Differentiable Mask Pruning} (DMP) for gradual pruning while training a network. Our method can be generalized to unstructured (i.e. weights) or structured (i.e. vector of parameters, filter, subnetwork) sparsity. We define DMP as follows.

Suppose $\mathbf X$ is the feature matrix. Let $\alpha~\in \mathbb{R_+}^{d}$ be a strictly positive scaling factor of dimension $d$ for a given prunable entity, $f$ be a scale sensitive differentiable function (i.e. $f(\alpha \odot \mathbf{X})~\neq f(\mathbf{X})$), $I(\alpha)$ be a mask function where 
\begin{equation}
I(\alpha) =  \begin{cases}
    1 & \text{if $|\alpha| >$ t},\\
    0 & \text{otherwise}.
  \end{cases}
  \label{eq:mask}
\end{equation} 
 and t be a small thresholding value. The intuition behind our approach is to to  replace $f(\mathbf{X})$ with  $g(\mathbf{X}) = f(\alpha \odot  I(\alpha) \odot \mathbf{X})$ and apply $\ell_1$-regularization on  $\alpha$  to intorduce sparsity on its corresponding prunable entity. Formally we propose to replace $J(\theta)$ with 
\begin{equation}
\begin{aligned}
J(\theta, \alpha) &=  C(\theta, \alpha) + \mathcal R(\theta) +  \lambda\sum_{l \in L}||\alpha_l||_1,
\label{eq:regularizer}
 \end{aligned}
\end{equation}
in which $C(\theta, \alpha)$ is the cost function,  $ \mathcal R(\theta)$ is regularizer, often an $\ell_2$ norm on weights. 
For \emph{filter pruning}, the prunable entities are the filters $\theta_l\in \mathbb{R}^{m \times n \times k \times k}$ in a convolution layer with $m$ filters, $n$ outputs kenel of size $k\times k$, in which $\alpha\in \mathbb{R_+}^m$ (i.e. one scaling factor per filter). We may consider $f(\mathbf{X}) = \mathrm{ReLU(BatchNorm(Conv}(\mathbf{X}, \theta_l))$. 

For \emph{subnetwork pruning}, the prunable entity is the subnetwork (Figure \ref{fig:resnet_block}), $\alpha \in \mathbb{R_+}$ (i.e. one scaling factor per subnetwork).  The output of the network is defined as $h(\mathbf X_1,\mathbf X_2) = f(\mathbf X_1) + \mathbf X_2$ where $f(\mathbf X_1)$ is the output of the subnetwork.

The mask function $I(\alpha)$ in  \eqref{eq:mask}, returns either 0 or 1 based on the value of $\alpha$ in comparison to a tiny positive  $t\approx 10^{-5}$. Within the loss function, we apply $\ell_1$ regularization on $\alpha$ to enforce sparsity over each of the pruning entities. The primary reason for using $I(\alpha)$ is to ensure that the pruning happens while training. The $\ell_1$ regularization may  not push the value to  zero numerically, and $I(\alpha)$ is used to correct this numerical error. Moreover, instead of using the regular $\ell_2$ regularization of the weights for preventing over-fitting, we make use of a modified $\ell_2$ regularization. The purpose is to ensure the weights that are already pruned do not contribute to the loss, unless the weights are rejuvenated through the differentiable masks. This is somehow a numerical correction to the elastic net penalty \cite{ZouHastie_ElasticNet_2005} motivated by \emph{garotte} correction \cite{Breiman_Garotte_1993} to an  $\ell_1$-regularized estimator.

The loss function can be defined therefore as
\begin{equation}
J(\theta,\alpha) = {C}(\theta,\alpha) + \lambda_1 \sum_{l \in L} ||\alpha_l||_1 + \lambda_2 \sum_{l \in L} ||I(\alpha_l) \theta||^2_2.
\end{equation}
The differentiable mask function is used in this case instead of a direct hard thresholding of the scaling factors such that the network learns to prune and un-prune filters. The network therefore learns to adjust its weights while maintaining its stability through the learning process. Therefore, as pruning is gradual as opposed to doing it all-at-once, the network has more flexibility to learn using a reduced number of parameters.

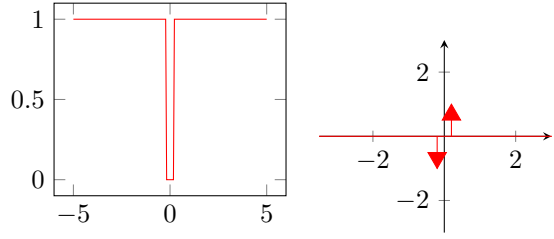
\begin{figure}
\centering
\begin{tikzpicture}
\begin{axis}[samples=200,scale = 0.45]

\addplot[color=red]{0.5*((sign(abs(x)-0.2))+1)};
\end{axis}

\end{tikzpicture}
\hspace{0.1in}
\begin{tikzpicture}
\makeatletter
\begin{axis}[axis lines=middle,xmin=-3.5,xmax=3,ymin=-3,ymax=3,scale = 0.45]
\addplot +[color=red,dirac] coordinates {(-0.2,-1) (0.2,1)};
\addplot [color=red] {0};
\end{axis}
\end{tikzpicture}

\caption{Original mask function (left) and its derivative (right).}
\label{fig:Original_thresh}
\end{figure}

\begin{figure}
\centering
\begin{tikzpicture}
\begin{axis}[samples=200, scale=0.45]
\addplot[color=red]{(1/(1+exp(-5*(abs(x)-0.2))))  + (5*(abs(x)-0.2))*((1/(1+exp(-5*(abs(x)-0.2))))*(1-((1/(1+exp(-5*(abs(x)-0.2)))))) )  };
\end{axis}
\end{tikzpicture}
\hspace{0.1in}
\begin{tikzpicture}
\makeatletter
\begin{axis}[samples=200,scale=0.45]
\addplot[color=red]{(5*(2-5*x*tanh(5*x/2))) / (1+cosh(5*x)) * sign(x) };
\end{axis}
\end{tikzpicture}

\caption{Approximated mask function (left) and its derivative (right).}
\label{fig:New_thresh}
\end{figure}
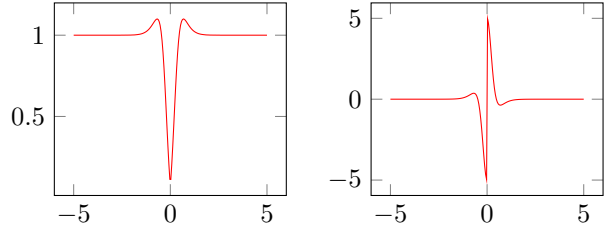

In Figure \ref{fig:Original_thresh}, the mask function is non-differentiable at two points and the overall gradient remains zero except at the threshold. Using the derivative of this hard-thresholding function during back-propagation, ends up freezing scaling factors. Approximating derivatives which provides gradients everywhere, mimics a straight-trough estimator, and is a common practice in quantization of deep networks. Having a continuous smooth approximation function close to the original instead of a plain straight-through estimators improves learning.  Therefore, we use the approximation of this mask function from the first derivative of foothill function \cite{belbahri_2019_foothill} and use its derivative (Figure \ref{fig:New_thresh}) in back-propagation to provide better gradient flow.
The equation below is the first derivative of the foothill function with unit shape parameter. We take the absolute value of this equation to obtain Figure \ref{fig:New_thresh}. For the backward pass, the derivative of the equation \eqref{foothill} is used as an approximation. The first derivative of the foothill function with unit shape parameter is 
\begin{equation} \label{foothill}
f(x,\beta) =  \tanh \left(\frac{\beta x}{2}\right) + \frac{1}{2}  \beta  x ~\mathrm{ sech}^2\left(\frac{\beta x}{2}\right),
\end{equation}
and its second derivative is
\begin{equation} \label{foothill-second-derivative}
    \frac{\partial f(x,\beta)}{\partial x} = \frac{1}{2} \beta~ \mathrm{sech}^2\left( \frac{\beta x}{2} \right) \left\{  2 - \beta x ~ \mathrm{tanh}\left( \frac{\beta x}{2} \right) \right\}
\end{equation}
see \cite{belbahri_2019_foothill} for more details.
Here is a brief highlight of our proposed method
\begin{enumerate}
\item Based on the value of $\alpha$, the prunable entity will either be retained or pruned as training proceeds.
\item 	Through the training of the network, the prunable entity can be recovered through the differentiable masks.
\item The mask function $I(\alpha)$ is used as a numerical error correction of $\alpha$ as $\ell_1$ penalty does not push the value exactly to zero. 
\item A smoother version of the mask function is used as an extension of the foothill function derivative to improve gradient flow and improve forward and backward match.
\end{enumerate}

\section{Pruning ratio control}
Our proposed technique controls the amount of filters to be pruned only  through the regularization constant $\lambda$ in \eqref{eq:regularizer}. To provide a more fine-grained control on the number of pruned parameters, we propose to solve the  constrained optimization
\begin{equation}
\begin{aligned}
   J^* &= \min_{\Theta}J(\theta) \\ & \text{ subject to } r(\theta_l) = c
\end{aligned}
\end{equation}
where $r(\theta_l)$ is the pruned ratio in model $\mathcal M$, given $c$ the target ratio. Motivated by the lasso, we recommend to minimize a soft-version which is practically easier 
\begin{equation}
\begin{aligned}
J(\theta, \alpha) 
=&  C(\theta, \alpha) + \lambda_1\sum_{l \in L}||\alpha_l||_1 + \lambda_2 \sum_{l \in L} ||I(\alpha_l) \theta||^2_2   \\ 
&+ \lambda_3\mathrm{max}\left\{0,\left(\frac{1}{\mathrm{K}}\sum_{l \in L}||I(\alpha_l)||_1 - c\right)\right\}.
\end{aligned}
\label{eq:allregs}
\end{equation}
Note that $K$ is the total number of filters in $\mathcal M$ and this additional regularizer is deactivated whenever it reaches the target ratio $1 - c$. 

Hyper-parameters of \eqref{eq:allregs} such as $\lambda_1$, $\lambda_2$, $\lambda_3$ need to be tuned to obtain  results that compete with state of the art. 
However, we recommend to use hyper-parameter optimization techniques to help  searching for the best values of these hyper-parameters if maximum accuracy is the target.

\section{Results}
We report results on a vision and a natural language processing task. For the vision task, we report the results on CIFAR-10 \cite{CIFAR10}, a dataset commonly used to benchmark methodological approaches. This vision task involves predicting an image's class between ten different categories. Similarly, to show the flexibility of our technique we use the IMDB \cite{IMDBSentiment} sentiment analysis dataset for sentiment classification, which involves classifying an input text as a positive or negative sentiment and  Penn Treebank dataset \cite{PTB_1993_Marcus},  a language model task for predicting the next word.

\textbf{CIFAR-10}: The dataset consists of 50k training images and 10k testing images that belong to one of the ten classes. The images are $32\times 32$ pixels with 3 channels for RGB. During training, we apply data augmentation by padding the images with 4 zeroes on each side then taking a random crop of $32 \times 32$ and randomly flipping them horizontally. During both training and testing, the inputs are normalized using $\mathrm{mean}=(0.4914, 0.4822, 0.4465)$ and $\mathrm{std}=(0.247, 0.243, 0.261)$ respectively. We first test our method and compare it with \emph{Network Slimming}  \cite{NetworkSlimming_Liu_2017} on the same VGG-19 architecture for filter pruning. To have a fair comparison, we use the same hyper-parameters and trained the network in a similar setting. We train the network for 160 epochs using the SGD optimizer and an initial learning set to 0.1 and divided by ten at epochs 80 and 120.
We perform subnetwork and filter pruning on ResNet-56 \cite{Resnet_He_2015}, a specialized residual network for CIFAR-10 dataset that consists of three downsampling stages of 9 subnetworks (called a ResNet basic block), with each subnetwork consisting of two $3\times 3$ convolution layers. We train the network for 170 epochs using the SGD optimizer with a momentum of $0.9$ and a starting learning rate of $0.1$, decayed by ten at epochs 80 and 160. The $\ell_2$-regularization is applied on the weights with $\lambda=10^{-4}$. The $\ell_1$-regularization is applied on the scaling factors with $\lambda=10^{-3}$. The threshold of the mask function is set to $10^{-3}$ for subnetwork pruning and to $10^{-4}$ for filter pruning. To show the versatility of DMP, we also show that it can be easily integrated with other compression techniques such as quantization \cite{Jacob_2017_QuantizationTflite}.

\textbf{IMDB sentiment classification}: The IMDB movie review dataset consists of 25k train and 25k test reviews that are either positive or negative sentiment. The network consists of an embedding layer, an LSTM layer (single stack) and a fully connected layer. We train a sentencepiece model for encoding the input dataset. The text is truncated if it exceeds 200 words. The vocabulary size used was 5k with embedding dimension 300 and hidden dimension 150. 

\textbf{PTB Language Model}: Penn Treebank dataset \cite{PTB_1993_Marcus}, also known as \emph{PTB dataset}, is widely used in machine learning of NLP research as a benchmark. The network is exactly the same as mentioned in the sentiment analysis task apart from the last fully connected layer that maps to the vocab size of 10k.

To elaborate the node pruning within the LSTM, we use the below conventional LSTM cell with the internal work flow as shown below:
\begin{eqnarray*}
f_t &=& \sigma (\mathbf{W_f}[h_{t-1},x_t] + b_f)\\
i_t &=& \sigma (\mathbf{W_i}[h_{t-1},x_t] + b_i)\\
g_t &=& \tanh (\mathbf{W_g}[h_{t-1},x_t] + b_g)\\
o_t &=& \sigma (\mathbf{W_o}[h_{t-1},x_t] + b_o)\\
c_t &=& f_t \odot c_{t-1} + i_t \odot g_t\\
h_t &=& o_t \odot \tanh(c_t)
\end{eqnarray*}

So the parameters within the cell are 4 weight matrices (with a stacking for hidden and input state). We apply the node pruning within the cell as follows:
\begin{eqnarray*}
f_t &=& I(\alpha_f) \odot \sigma \Big(\alpha_f \odot  (\mathbf{W_f}[h_{t-1},x_t] + b_f)\Big)\\
i_t &=& I(\alpha_i) \odot \sigma \Big(\alpha_i \odot  (\mathbf{W_i}[h_{t-1},x_t] + b_i)\Big)\\
g_t &=& I(\alpha_g) \odot \tanh \Big(\alpha_g \odot  (\mathbf{W_g}[h_{t-1},x_t] + b_g)\Big)\\
o_t &=& I(\alpha_o) \odot \sigma \Big( \alpha_o \odot  (\mathbf{W_o}[h_{t-1},x_t] + b_o)\Big)
\end{eqnarray*}

For IMDB sentiment analysis, the models were trained for 50 epochs with an initial learning rate of $10^{-4}$ and divided by 10 at epoch 20. The $\ell_2$ regularization of $10^{-4}$  was used in all experiments. The training technique for the language model is chosen from \cite{mikolov_2018_RNN}, using a medium LSTM model with hidden Dimension 650. 
\subsection{Filter Pruning}
For filter pruning, we show flexibility of our method for two sets of models, VGG-19 and ResNet-56. Compared to Network Slimming (NS) \cite{NetworkSlimming_Liu_2017} using one iteration, our Differentiable Mask Pruning (DMP) is able to prune more parameters $\approx 8\%$ and floating-point operations $\approx 30\%$ with only a negligible increase in test error $+0.13$ (Table \ref{VGG19_CIFAR10}). After five iterations of prune-then-finetune, NS obtains similar compression ratio as DMP at the cost of an increase in error of $\approx 1.50\%$. We can clearly see that the network benefits from being pruned while training as it only requires one iteration to achieve a similar compression ratio compared to five iterations of NS. Note that we did not fine-tune the initial hyper-parameters when applying our method to have a side-by-side comparison, but with proper tuning we expect to close the gap between our pruned model and the initial unpruned baseline. We further confirm this intuition on ResNet-56 (Table \ref{RESNET56_CIFAR10_Filter}) for different set of $\lambda_1$ values. For $\lambda_1=10^{-4}$ to $\lambda_1=10^{-3}$, there is a smooth transition for the number of pruned filters ranging from pruning ratios of $0.12$ to $0.78$, while the accuracy drop is relatively low compared to the baseline. There is versatility of DMP to combine pruning with 8-bit quantization. We also evaluate our proposed control of pruning ratio regularizer on ResNet-56 with different target ratio ${0.4, 0.6, 0.8}$ and compare the actual target ratio after training. We observe that number of remaining parameters are within the range of this target ratio constraint.

\begin{table}[h]
\centering
\begin{footnotesize}
\begin{tabular}{| c  c  c  c |}

\hline
Method  & Test error & Pruned Params & Pruned Flops \\ 
\hline
Unpruned  & 6.01  & - & - \\
\hline
Network &&&\\
Slimming & 6.20  & 0.88 & 0.51  \\
\hline
DMP  & 6.33  & 0.96 & 0.80  \\
\hline
\end{tabular}
\vspace{3px}
\caption{ VGG-19 architecture run on CIFAR-10 to prune filters.}
\label{VGG19_CIFAR10}
\end{footnotesize}
\end{table}

\begin{table}[h]
\centering
\begin{footnotesize}
\begin{tabular}{| c  c  c   c  c|}

\hline
Method & $\lambda_1$ & Test   &  Pruned filters  & Ratio \\ 
 & $\times 10^{-4}$ &   Error   & out of 2032 & $\times 100$\\
\hline
Unpruned  & - & 6.53 & 0 & 0 \\
\hline
DMP & $1$ & 6.81 &   264 & 12 \\
\hline
DMP & $5$ & 8.48 &  1227 & 60 \\
\hline
DMP & $10$ & 9.50 & 1599 & 78 \\
\hline

\end{tabular}
\vspace{3px}
\caption{ResNet-56 architecture run on CIFAR-10, to prune filters with different regularization constant $\lambda_1$  \eqref{eq:allregs}.}
\label{RESNET56_CIFAR10_Filter}
\end{footnotesize}
\end{table}

\begin{figure*}
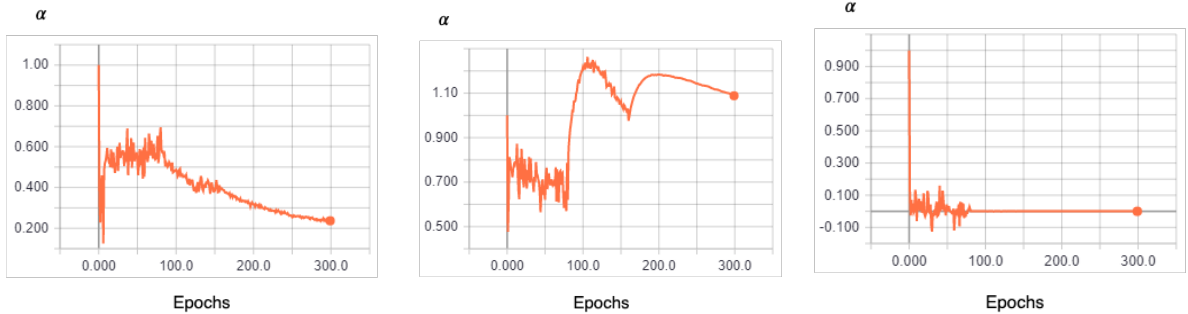

  \centering
  \includegraphics[width=0.32\linewidth]{blockprune_alpha_1.png}
  \includegraphics[width=0.32\linewidth]{blockprune_alpha_2.png}
  \includegraphics[width=0.32\linewidth]{blockprune_alpha_3.png}
  \caption{The training process of 3 scaling factors for subnetwork pruning. As one scaling factor gets pruned, another scaling factor compensates for the instability caused and increases in magnitude. The rightmost figure also shows the scaling factor rejuvenating throughout the training process and finally pruned at epoch 80.}
  \label{fig:subnetwork_Scalingfactor}
\end{figure*}

\subsection{Subnetwork pruning}
Subnetwork pruning can be seen as a generalization of architecture search, where our DMP method is similar to architecture search using DARTS \cite{Liu_2018_DARTS}. We evaluate the versatility of DMP on this task with ResNet-56 that consists of 27 subnetworks (ResNet basic blocks). We observe a smooth transition in the number of pruned subnetworks ranging from $5/27$ to $17/27$ while having a relatively low drop of $\sim 1.3\%$ in accuracy (Table \ref{RESNET56_CIFAR10_Subnetwork}). The resulting architecture for the experiment that pruned $17/27$ blocks is 3, 2 and 5 subnetworks in the first, second and last stage respectively. A snapshot of scaling factors during the learning process is shown in Figure \ref{fig:subnetwork_Scalingfactor}.

\begin{table}[h]
\centering
\begin{footnotesize}
\begin{tabular}{| c   c   c  c  c c | }

\hline
Method & $\lambda_1$ & Test   &   Pruned subnets &  Pruned filters & Ratio \\ 
 & $\times 10^{-3}$& Error& out of 27& out of 2032 &$\times 100$ \\
\hline
Unpruned  & - & 6.53 & 0 &  0 & 0\\
\hline
DMP & $1$ & 7.10 &  5 & 320 & 15.70 \\
\hline
DMP & $5$ & 7.78  & 12 & 928& 45.60 \\
\hline
DMP & $10$ & 8.34  & 17 & 1152& 56.60 \\
\hline
\end{tabular}
\vspace{3px}
\caption{ResNet-56 run on CIFAR-10 to prune subnetworks with different regularization constant $\lambda_1$.}
\label{RESNET56_CIFAR10_Subnetwork}
\end{footnotesize}
\end{table}

\subsection{Filter versus sub-network pruning}
For the ResNet56 architecture, we show the results of both filter (Table \ref{RESNET56_CIFAR10_Filter}) and subnetwork (Table \ref{RESNET56_CIFAR10_Subnetwork}) pruning. The results indicate the flexibility and robustness of our technique with similar pruning results on filter and subnetwork pruning done independently on the same architectures.   
However, while comparing the effects of pruning and the overall accuracy, empirically, subnetwork pruning provides a slightly better trade-off between pruning and accuracy drop. We speculate that pruning the whole subnetwork reduces the overall noise in the network and hence acts as a better regularizer.

\subsection{Node Pruning}

For the node pruning technique within the LSTM, we compare our results to the baseline  on the LSTM network and the results obtained show the versatility of the methodology on a text dataset on both sentiment analysis and language model tasks, as shown in Table~\ref{IMDB} and Table~\ref{Langauge_Model}.

\begin{table}[h]
\centering
\begin{footnotesize}
\begin{tabular}{| c  c  c  c   c | }

\hline
Method  & $\lambda_1$ & Test  & Pruned nodes & Ratio   \\ 
& $\times 10^{-2}$ & Error & out of 600 & $\times 100$ \\
\hline
Unpruned  & - &13.43  & 0& 0  \\
\hline
DMP & $1$  & 13.61  & 540 & 90.0   \\
\hline
DMP & $5$  & 13.78  & 543 & 90.5   \\
\hline
DMP & $10$  & 13.94  & 577 & 96.0   \\
\hline
\end{tabular}
\vspace{3px}
\caption{IMDB Sentiment Analysis using recurrent networks to prune LSTM nodes.}
\label{IMDB}
\end{footnotesize}
\end{table}

\begin{table}[h]
\centering
\begin{footnotesize}
\begin{tabular}{| c  c  c  c   c | }

\hline
Method  & $\lambda_1$ & Test   & Pruned nodes  & Ratio   \\ 
& $\times 10^{-7}$ & perplexity & out of 5200 & $\times 100$ \\
\hline
Unpruned  & - &84.90  & 0 & 0  \\
\hline
DMP & $1$  & 85.10  & 524 & 10   \\
\hline
DMP & $5$  & 85.02  & 672 & 12    \\
\hline
DMP & $10$  & 86.10  & 940 & 18   \\
\hline
\end{tabular}
\vspace{3px}
\caption{Penn Tree Bank Dataset Language model using recurrent networks to prune LSTM nodes.}
\label{Langauge_Model}
\end{footnotesize}
\end{table}

Overall, DMP obtains competitive results on node, filter and subnetwork pruning on two different tasks applied on a  variety of neural network architectures and showcases the potential and simplicity of the technique.

\section{Conclusion}
We introduced DMP, a new technique that extends pruning on two directions: structured and unstructured.  DMP induces sparsity that can be easily extended to prune  weights, nodes, vectors, filters and sub-networks. The main shortcoming of pruning is to train the network properly with fewer parameters. We proposed to improve the training procedure by approximating the hard threshold gradient, and updating back-propagation accordingly. DMP provides the flexibility to recover pruned weights and improves the learning capacity of the pruned network during training. Additionally, DMP even shows its versatility through the ease of integration with quantization. Here, we only focused on simple vision and text tasks, but our initial experiments show promising performance on  larger architectures and more complex data. If pruning entity is a sub-network, DMP can be regarded as a differentiable architecture search method, while spanning always on architectures with  lower complexity.

\section*{Acknowledgement}
We appreciate fruitful technical discussions with Huawei Cloud Core Shanghai  Gang  Chi  and  Pengcheng  Tang, and  Huawei Intelligent Computing Ascend Computing Toronto Hamed Mousazadeh and Gordon Deng.  We  also thank Yanhui Geng, Steven Yuan, and Yan Zhou for their support throughout the project.

\bibliographystyle{IEEEtranBST/IEEEtran}
\bibliography{pruning}

\end{document}